\documentclass[letterpaper, 10 pt, conference]{ieeeconf}  

\IEEEoverridecommandlockouts                              

\usepackage{cite}
\usepackage{amsmath,amssymb,amsfonts}
\usepackage{algorithmic}
\usepackage{graphicx}
\usepackage{float}
\usepackage{dblfloatfix}
\usepackage{textcomp}
\usepackage[dvipsnames]{xcolor}
\usepackage{siunitx}
\usepackage{xfrac}
\usepackage{subcaption}
\usepackage{float}
\usepackage{multirow}
\usepackage{hyperref} 
\hypersetup{hidelinks}
\usepackage{cleveref}
\usepackage[utf8]{inputenc}

\title{\LARGE \bf
	\mbox{3-D}imensional Sonic Phase-invariant Echo Localization
}

\author{Christopher Hahne$^{1}$
	\thanks{*This project is funded by the Hasler Foundation under number 22027.}
	\thanks{$^{1}$Christopher Hahne is affiliated with the Artificial Intelligence on Medical Imaging group at the ARTORG Center, University of Bern, 3008 Bern, Switzerland
		{\tt\small christopher.hahne@unibe.ch}}%
}

\begin{document}
		
	\maketitle
	\thispagestyle{empty}
	\pagestyle{empty}
	\begin{abstract} 
		Parallax and Time-of-Flight (ToF) are often \mbox{regarded} as complementary in robotic vision \if. as complementary robotic vision techniques. \fi
		where various light and weather conditions remain challenges for advanced \mbox{camera-based} \mbox{3-Dimensional} \mbox{(3-D)} reconstruction. 
		To this end, this paper establishes Parallax among Corresponding Echoes (PaCE) to triangulate acoustic ToF pulses from arbitrary sensor positions in \mbox{3-D} space for the first time. %
		This is achieved through a novel \mbox{round-trip} reflection model that pinpoints targets at the intersection of ellipsoids, which are spanned by sensor locations and detected arrival times. 
		\mbox{Inter-channel} echo association becomes a crucial prerequisite for target detection and is learned from feature similarity obtained by a stack of Siamese Multi-Layer Perceptrons (MLPs). 
		The PaCE algorithm enables phase-invariant \mbox{3-D} object localization from only 1 isotropic emitter and at least 3 ToF receivers with relaxed sensor position constraints. 
		Experiments are conducted with airborne ultrasound sensor hardware and back this hypothesis with quantitative results. 
		The code and data are available at \texttt{{\color{NavyBlue}\textmd{\url{{https://github.com/hahnec/spiel}}}}}.
	\end{abstract}
	
	\section{Introduction}
	Certain animal species that move in \mbox{\mbox{3-D}imensional} \mbox{(3-D)} space perceive surroundings by pulses emitted and bounced off from obstacles. Can tomorrow's robots do likewise? \par
	Over recent decades, \mbox{3-D} computer vision has been dominated by stereoscopic parallax and Time-of-Flight \mbox{(ToF)} sensing. Depth perception in the light spectrum is a \mbox{well-studied} subject adopted by Simultaneous Localization and Mapping (SLAM) to help robots navigate. 
	However, varying weather (e.g., fog, rain, etc.) or severe lighting conditions impair computational imaging. Only a little attention has thus far been given to \mbox{3-D} reconstruction from detectors working at other wavelengths. 
	To address these challenges, this paper introduces Parallax among Corresponding Echoes (PaCE) as a depth-sensing hybrid incorporating triangulation and ToF concepts at a geometric level. To the best of the author's knowledge, this is the first systematic feasibility study on \mbox{3-D} object localization from parallax-based ToF being invariant of the frequency and phase. 
	\par
	\begin{figure}[!t]
        \vspace{2mm}
		\centering
		\begin{minipage}{.62\linewidth}
			\includegraphics[width=\textwidth]{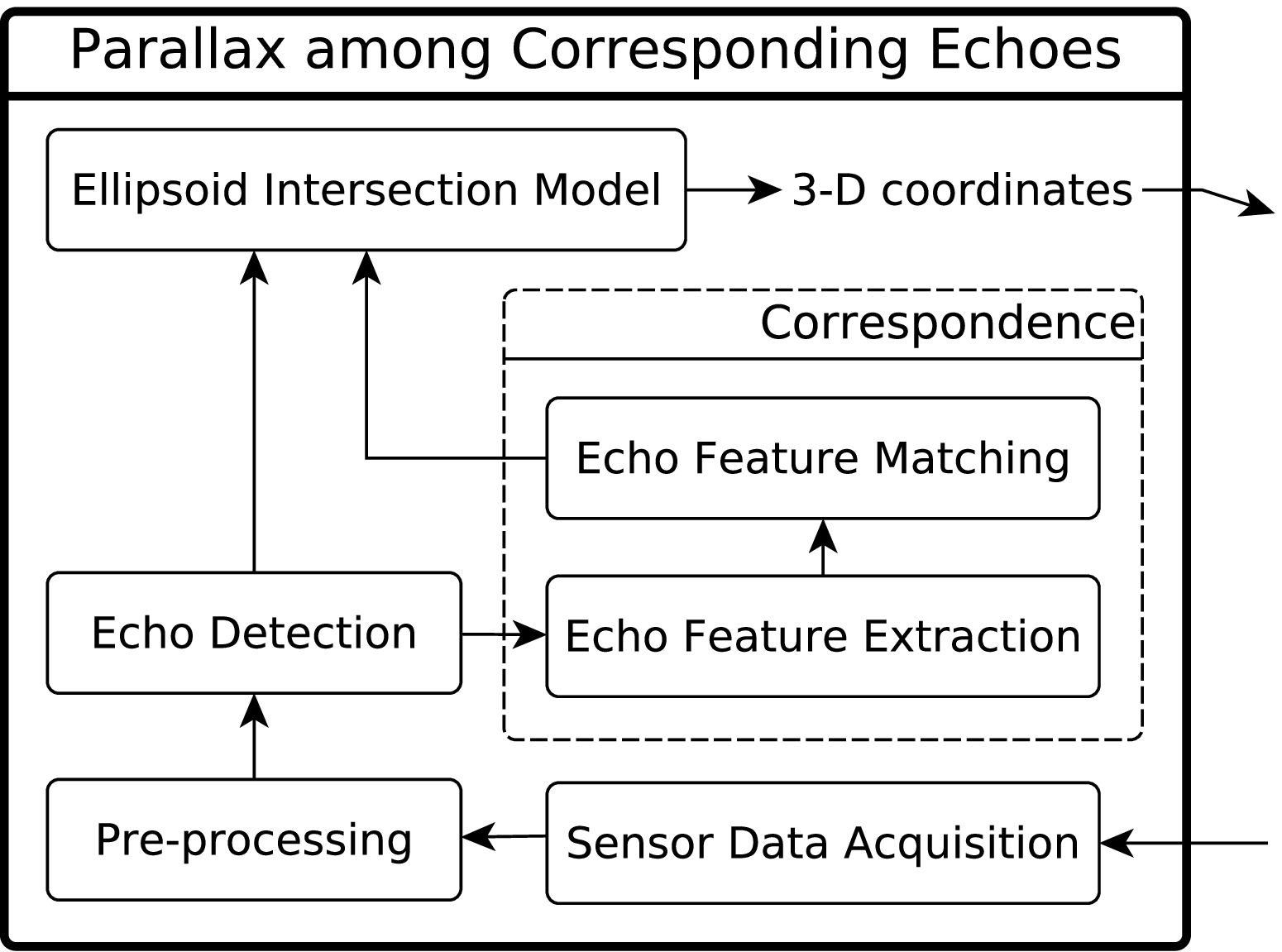}
		\end{minipage}
		\hfill
		\begin{minipage}{.35\linewidth}
			\includegraphics[trim=700 130 520 530,clip=true,width=\textwidth]{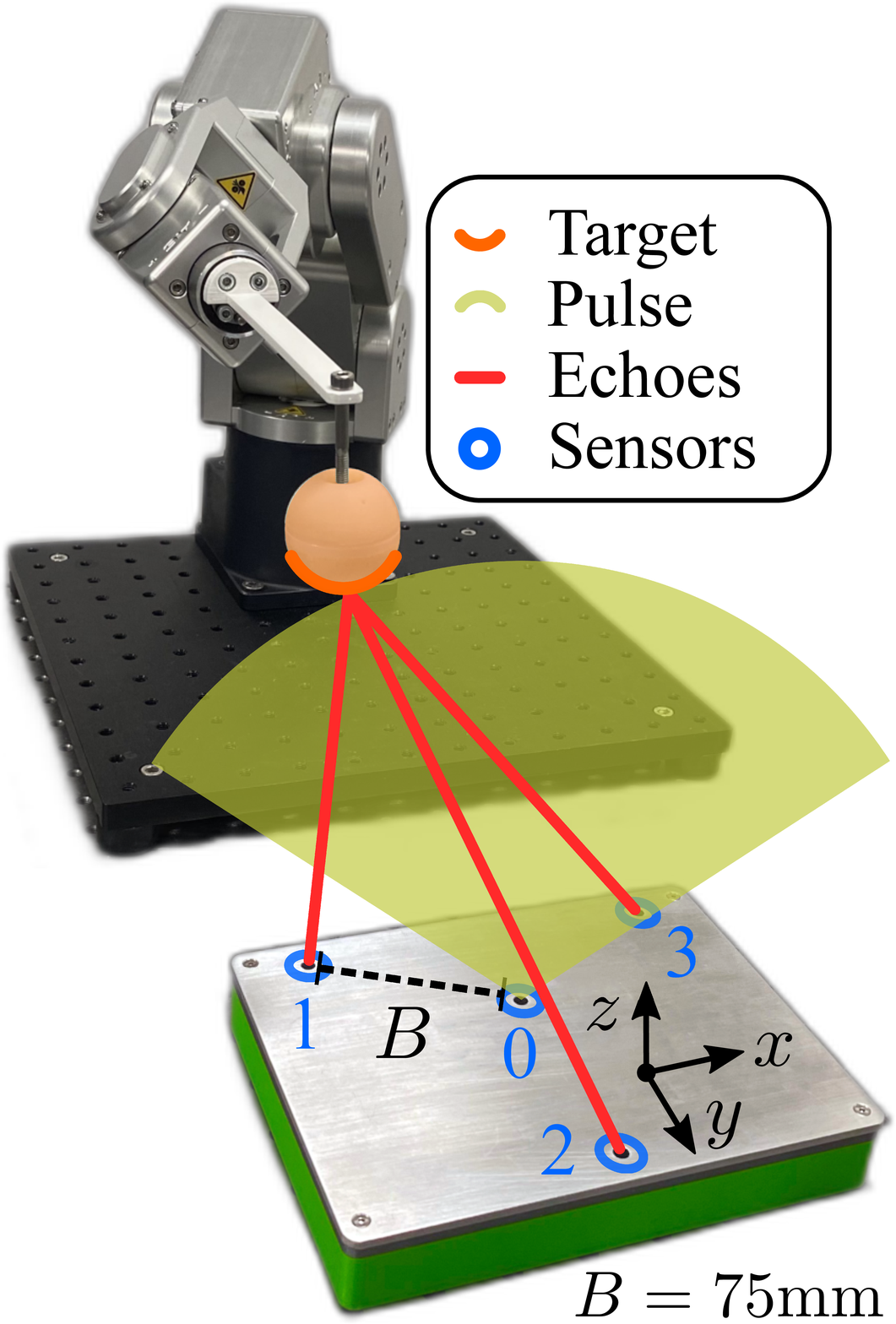}
		\end{minipage}
		\caption{Robotic navigation demands \mbox{3-D} sensing in challenging environments. This study proposes Parallax among Corresponding Echoes (PaCE) as a novel framework (left) to triangulate echoes in \mbox{3-D} space (right). 
		\label{fig:top}}
	\end{figure}
	%
	Early research on sonar-based echolocation for mobile robots retrieved object points in \mbox{2-D} utilizing ellipse intersections~\cite{Peremans:1993,Wijk:2000,Bank:2007}. There is also a considerable amount of patents that claim \mbox{2-D} ellipse intersections as a localization method lately~\cite{Mitsubishi:2016,Toposens:2016,Valeo:2019}. Several studies attempted to mimic a bat's acoustic perception by modelling ears with two microphones and inferring obstacle locations using spherical coordinates~\cite{Reijniers:2007,Schillebeeckx:2011, Eliakim:2018, Christensen:2020}. A recent breakthrough in \mbox{3-D} reconstruction from audible acoustics is based on deep learning architectures trained by camera-based depth maps without physical modelling~\cite{Christensen:2020,Tracy:2021,Parida:2021,Irie:22}.\par
	%
	The existing literature and inventions disregarded the potential of \mbox{3-D} localization from intersecting ellipsoids and a preceding echo association. Here, echo association refers to the correct assignment of echoes to actual targets. Opposed to phased arrays, where transducers are separated by a multiple of the wavelength, the presented echo correspondence and geometric localization model fill this gap by relaxing sensor and position constraints in \mbox{3-D} sonar tracking. This study hypothesizes that phase correlation methods can be substituted with machine learning algorithms to gain confidence in the localization and achieve more flexibility on the hardware and application side. In particular, the novel ellipsoid intersection model proposed in this work is considered the most generic solution for \mbox{3-D} localization from at least 3 detectors and only one emitter without constraints on their positions in space. A European patent application covering PaCE has been filed by the University of Bern~\cite{pace:2022}. 
	\par
	This paper begins with a literature overview in Section~\ref{sec:related}. The ellipsoid intersection model is introduced in Section~\ref{sec:ellipsoid}, establishing the need for echo association in Section~\ref{sec:correspondence}, where further details on novel algorithmic aspects are provided. The experimental work in Section~\ref{sec:experiment} shows results rendered by PaCE as a proof-of-concept. Section~\ref{sec:conclusions} concludes while reflecting on the framework's potential and prospects. 
	
	\section{Related Work}
	\label{sec:related}
    %
	Technology families related to PaCE are Phased-Arrays (PAs), stereoscopic vision cameras, and Real-Time Locating Systems (RTLS) using Time Difference of Arrival (TDoA). PaCE is an active ToF-based triangulation method and substantially different from prior work. Its localization scheme is phase-invariant, detectors can be arbitrarily positioned, and targets are not carrying a tag or beacon unit. Also, PAs generally comprise a large number of transducers, which - from the author's standpoint - is considered a redundancy. Thus, the motivation of this study originates from the idea that the transducer number requirement in PAs can be traded for computational efforts while aiming for comparable performance. \par
	An initial attempt for sonar (sound navigation and ranging) with a sparse number of sensors \if can be traced\fi dates back to Peremans~\textit{et al.}~\cite{Peremans:1993}, who had striven to locate objects along a 2-D plane for robotic navigation. A follow-up study by Wijk and Christensen extended this triangulation-based approach by fusing measurements from different points in time for 2-D mobile robot indoor pose tracking~\cite{Wijk:2000}. Bank and Kämpke generalized this 2-D triangulation by proposing tangential regression for ellipse intersections and built a robot equipped with an array of transducers to reconstruct a high-resolution 2-D map of surroundings~\cite{Bank:2007}. Notably, intersection methods were reported as part of trilateration in the radar imaging field~\cite{Ahmad:2006, Mirbach:2011}. For instance, Malanowski and Kulpa explored 3-D target localization based on 3 transmitters and a receiver for multi-static radar in aeronautics~\cite{malanowski:2012}. In the same years, the group led by Peremans published an avid study working towards \mbox{3-D} localization based on the direction of spectral cues from two bat-shaped outer ears in an attempt to mimic bat perception~\cite{Schillebeeckx:2011}.  Kuc and Kuc investigated echolocation as an orientation assistance for blind people~\cite{kuc:2016}. More recently, a 2-D localization system for pen or finger tracking was proposed by Juan and Hu~\cite{Chung-Wei:21}, who employed multiple ultrasonic sensors in conjunction with Newton–Raphson optimization and Kalman filtering to recover 2-D positions at standard deviations below 1~\si{\centi\meter}. \par
	An emerging related research field concerns learned acoustic \mbox{3-D} reconstruction in the audible frequency range~\cite{Christensen:2020,Tracy:2021,Parida:2021,Irie:22}. In an experimental study, Generative Adversarial Networks (GANs) are trained from audible sweep chirps with the goal of recreating stereo depth maps from only a speaker and at least two consumer microphones~\cite{Christensen:2020,Tracy:2021,Parida:2021,Irie:22}. However, end-to-end supervised learning of acoustic \mbox{3-D} reconstruction is in an early research stage and faces challenges from potential biases implicit to the datasets, domain gap and disturbances from other audible sound sources, as commonly encountered in industrial environments.\par
	Over the last decade, peers have patented ultrasound transducer setups for object localization based on 2-D ellipse intersection models~\cite{Mitsubishi:2016,Toposens:2016,Valeo:2019}. The start-up Toposens GmbH has taken up from there with commercial products targeting industrial environments~\cite{Toposens:2016}. Their solution employs two consecutive ellipse intersections and requires orthogonally arranged transducers and a phase correlation method~\cite{Toposens:2016}. Future trends on ultrasound systems indicate a surge of collaborative and autonomous robotics in medical applications demanding effective localization capabilities~\cite{Haxthausen:2021}. \par
	The herein demonstrated framework complements prior work by enabling round-trip 3-D tracking from at least 3 sensors using a novel ellipsoid intersection model. This method is distinct in that it entirely relieves sensor position constraints. In particular, the proposed framework enables all sensors to be arbitrarily located or even move freely in 3-D space as long as their current position is known. Further, the presented method is invariant of the phase signal, distinguishing it from methods that rely on beamforming or matched filters. Instead, echo correspondence is recognized as an ambiguity problem that may potentially yield false object locations. Correct matching is addressed by training a Siamese MLP stack with features from Multimodal Exponentially Modified Gaussian (MEMG) distributions to overcome this ambiguity. Thereby, the proposed framework avoids the need for mechanical extensions such as waveguide or baffle designs and prevents related artefacts (e.g., cross-talk). To the best of the author's knowledge, this work presents the most generic model-based 3-D target localization scheme for a sparse number of arbitrarily located sensors. %
	\section{Ellipsoid Intersection}
	\label{sec:ellipsoid}
	\subsection{Echo Detection}
	In classical 1-dimensional (1-D) range finding, the transmitting and receiving transducers are identical, which implies that the outward and return travel paths coincide. In this special case, scalar distances $s^\star_k$ from $k\in\{1,2,\dots,K\}$ echoes can be readily obtained by
	\begin{align}
	s^\star_k = c_s\frac{t^\star_k}{2} \, , \, \text{where} \,\, t^\star_k = \{ t_i | t_i \in \mathbb{R}^T \wedge \nabla_t \left|\mathcal{H}\left[y_n(t_i)\right]\right| > \tau \} \label{eq:distance}
	\end{align}
	where $y_n(t_i)$ denotes the captured amplitude data from sensor $n$, $c_s$ is the propagation velocity and the divider accounts for the equidistant forward and backward travel paths. Each time sample $t_i\in\mathbb{R}^T$, with a total number of $T$, qualifies to be a detected Time-of-Arrival (ToA) denoted by $t^\star_k$ once the gradient of the Hilbert-transformed magnitude $\nabla_t \left|\mathcal{H}\left[y_n(t_i)\right]\right|$ surpasses a threshold $\tau$. Note that such a single sensor setup generally yields radial distances $s^\star_k$ making accurate directional information retrieval of the surrounding targets intractable. 	
	
	\subsection{Ellipsoid Surface Geometry}
\begin{figure}[ht]
		\centering
        \vspace{2mm}
		\includegraphics[width=.95\linewidth]{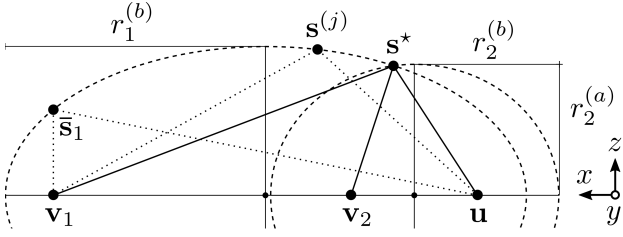}
		\caption{\textbf{Cross-sectional ellipsoid intersection} showing transmitter $\mathbf{u}$ and receivers $\mathbf{v}_n$ with radii $r^{(a)}_n$, $r^{(b)}_n$. Surface points $\mathbf{\bar{s}}_n$ and location candidates $\mathbf{s}^{(j)}$ span triangles (dotted lines) on the continuous ellipsoidal solution space (dashed curves).\label{fig:ellipse}}
	\end{figure}

	Pinpointing a \mbox{3-D} landmark generally involves advanced geometric modelling. Using a receiver and a transmitter in a so-called round-trip setup, ToA detection yields $t^\star_k$, whereas outward and return travel paths may be non-equidistant, i.e., \eqref{eq:distance} does not hold. This is because a distinct receiver position causes the travel direction to change after target reflection spanning a triangle between a possible target position \mbox{$\mathbf{\bar{s}}_1$}, transmitter \mbox{$\mathbf{u}\in\mathbb{R}^3$} and receiver \mbox{$\mathbf{v}_1\in\mathbb{R}^3$} (see Fig.~\ref{fig:ellipse}). %
	This triangle has its roots in the parallax concept, where an object point is observed from at least 2 different viewpoints~\cite{hahne2017baseline}. The vector between $\mathbf{u}$ and $\mathbf{v}_n$ can be regarded as the \textit{baseline}, and while this is given, the triangle's side lengths (i.e., travel paths) remain unknown in the single receiver case. Here, all travel path candidates form triangles with equal circumferences fixed at the baseline. Closer inspection of Fig.~\ref{fig:ellipse} reveals that feasible object positions $\mathbf{\bar{s}}_n\in\mathbb{R}^3$ yield an infinite set of solutions located on an ellipse for a \mbox{2-D} plane, and - when extended to \mbox{3-D} space - this solution set is represented by an ellipsoid. The surface of an ellipsoid thus reflects potential target locations in a \mbox{3-D} round-trip scenario comprising a single transmitter and receiver. %
	Adding a second receiver capturing a ToA from the same target spans a second ellipsoid that intersects with the first ellipsoid along a curve, carrying a subset of solution points and thus narrowing down the position candidate set. Only by introducing a third receiver and its respective ellipsoid, the target's \mbox{3-D} location ambiguity can be resolved as the surface curve and third ellipsoid exhibit at an intersection point in the send direction. It is mathematically demonstrated hereafter that a group of detected echoes $t^\star_{n,k}$ reflected from the same object and captured by $N\geq3$ sensors enables retrieval of the target position that resides on $N$ ellipsoid surfaces. \par
	Let any point \mbox{$\mathbf{\bar{s}}_n=\left[\bar{x}_n, \bar{y}_n, \bar{z}_n\right]^{\intercal}$} lie on the surface of ellipsoid~$n$ if $Q\left(\mathbf{\bar{s}}_n, \mathbf{r}_n\right) = 0$, which is given by
	\begin{align}
	Q\left(\mathbf{\bar{s}}_n, \mathbf{r}_n\right) = 
	\left(\frac{\bar{x}_n}{r_n^{(a)}}\right)^2 + \left(\frac{\bar{y}_n}{r_n^{(b)}}\right)^2 + \left(\frac{\bar{z}_n}{r_n^{(c)}}\right)^2-1 \label{eq:ellipsoid}
	\end{align}
	where $\mathbf{r}_n=[r^{(a)}_n, r^{(b)}_n, r^{(c)}_n]^\intercal$. 
	is a radii vector. It consists of a major axis $r^{(b)}_n$ drawn from $t^\star_{n,k}$ which is given by
	\begin{align}
	r^{(b)}_n = \frac{t^\star_{n,k}}{2}
	\end{align}
	and the minor axes $r^{(a)}_n = r^{(c)}_n$ are obtained by
	\begin{align}
	r^{(a)}_n = r^{(c)}_n = \frac{1}{2} \, \sqrt{\left(t^\star_{n,k}\right)^2-\lVert\mathbf{u}-\mathbf{v}_n\rVert^2_2}
	\end{align}
	with $\mathbf{u} \in \mathbb{R}^{3\times1}$ as the transmitter and $\mathbf{v}_n \in \mathbb{R}^{3 \times 1}$ for receiver positions found at the focal points of each ellipsoid. In fact, note that the two minor axes span oblate spheroids. The above definitions are only valid for those ellipsoids whose center resides on the coordinate origin and whose axes $\mathbf{r}_n$ are in line with the coordinate axes.
	Generally, a transducer ellipsoid may be displaced and arbitrarily oriented. To account for that, global space coordinates \mbox{$\mathbf{s}=\left[x, y, z\right]^{\intercal}$} are translated by an ellipsoid center \mbox{$\mathbf{c}_n=[\hat{x}_n, \hat{y}_n, \hat{z}_n]^\intercal$} and mapped onto its surface by a rotation matrix \mbox{$\mathbf{R}_n \in \text{SO}(3)$} so that
	\begin{align}
	\mathbf{\bar{s}}_n = \mathbf{R}_n^{\intercal}\left(\mathbf{s}-\mathbf{c}_n\right) \label{eq:transform}
	\end{align}
	which makes use of $\mathbf{R}_n^{\intercal}=\mathbf{R}_n^{-1}$ as a rotation matrix property.
	
	\subsection{Intersection via Root-finding}
	According to the aforementioned geometric definitions, a potential target ideally resides on the surface of at least 3 ellipsoid bodies. In a mathematical sense, this statement holds true for a point $\mathbf{s}^\star \in \mathbb{R}^3$ that satisfies
	\begin{align}
	Q\left(\mathbf{R}^{\intercal}_n(\mathbf{s}^\star-\mathbf{c}_n), \mathbf{r}_n\right)=0 \,\, , \quad \forall n \label{eq:zero-crossing}
	\end{align}
	by plugging \eqref{eq:transform} into \eqref{eq:ellipsoid}. Consequently, solving for $\mathbf{s}^\star$ breaks down to classical root-finding, so employing a multivariate Gradient Descent (GD) method is sufficient here. The GD update at iteration $j$ with step size $\gamma$ reads
	\begin{align}
	\mathbf{s}^{(j+1)} = \mathbf{s}^{(j)} - \gamma\mathbf{J}^{-1}\mathbf{f}
	\end{align}
	where the ellipsoid function vector $\mathbf{f} \in \mathbb{R}^{N\times1}$ is given by
	\begin{align}
	\mathbf{f} = 
		\begin{bmatrix} 
		Q\left(\mathbf{\bar{s}}_1^{(j)}, \mathbf{r}_1\right), 
		Q\left(\mathbf{\bar{s}}_2^{(j)}, \mathbf{r}_2\right), 
		\dots, 
		Q\left(\mathbf{\bar{s}}_N^{(j)}, \mathbf{r}_N\right) 
		\end{bmatrix}^\intercal
	\end{align}
	with $\mathbf{\bar{s}}_n^{(j)} = \mathbf{R}_n^{\intercal}\left(\mathbf{s}^{(j)}-\mathbf{c}_n\right)$. The Jacobian $\mathbf{J} \in \mathbb{R}^{N\times3}$ w.r.t. $\mathbf{s}^{(j)}$ is composed of analytical partial derivatives obtained by
	\begin{align}
	\mathbf{J} = 
	\begin{bmatrix}
	\partial_{x} Q(\mathbf{\bar{s}}_1^{(j)}, \mathbf{r}_1) & 
	\partial_{y} Q(\mathbf{\bar{s}}_1^{(j)}, \mathbf{r}_1) & 
	\partial_{z} Q(\mathbf{\bar{s}}_1^{(j)}, \mathbf{r}_1) \\[.8em]
	\partial_{x} Q(\mathbf{\bar{s}}_2^{(j)}, \mathbf{r}_2) & 
	\partial_{y} Q(\mathbf{\bar{s}}_2^{(j)}, \mathbf{r}_2) & 
	\partial_{z} Q(\mathbf{\bar{s}}_2^{(j)}, \mathbf{r}_2) \\[.8em]
	\vdots & 
	\vdots & 
	\vdots \\[.8em]
	\partial_{x} Q(\mathbf{\bar{s}}_N^{(j)}, \mathbf{r}_N) & 
	\partial_{y} Q(\mathbf{\bar{s}}_N^{(j)}, \mathbf{r}_N) & 
	\partial_{z} Q(\mathbf{\bar{s}}_N^{(j)}, \mathbf{r}_N) \\
	\end{bmatrix}
	\end{align}
	which are computed for each iteration $j$ until convergence. 
	The estimated location $\mathbf{s}^\star=\left[x^\star,y^\star,z^\star\right]^\intercal$ is selected via
	\begin{align}
		\mathbf{s}^\star=\underset{\mathbf{s}^{(j)}}{\operatorname{arg\,min}} \, \left\{\sum_{n=1}^N Q\left(\mathbf{R}^{\intercal}_n(\mathbf{s}^{(j)}-\mathbf{c}_n), \mathbf{r}_n\right)\right\}
	\end{align}
	considering $N$ ellipsoid surfaces.
	
	\section{Echo Correspondence}
	\label{sec:correspondence}
\begin{figure}[ht]
    	\centering
        \vspace{2mm}
    	\begin{minipage}{\linewidth}
    		\includegraphics[width=\textwidth]{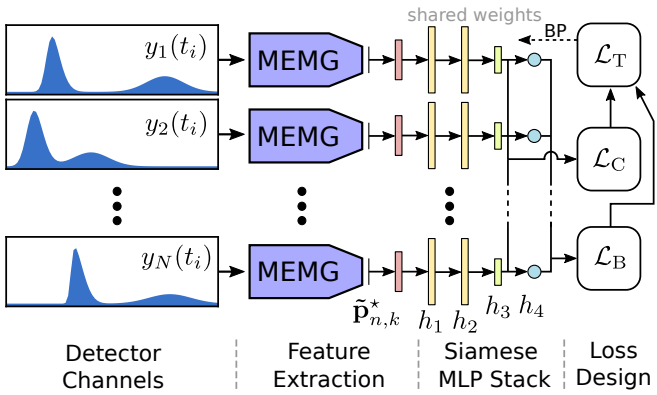}
    	\end{minipage}
    	\caption{\textbf{Echo correspondence architecture} where each \mbox{pre-processed} detector channel data $y_n(t_i)$ undergoes MEMG optimization~\cite{Hahne:2022}, providing features fed into a Siamese MLP stack. The overall training loss $\mathcal{L}_{\text{T}}$ aggregates the Binary Cross-Entropy (BCE) loss $\mathcal{L}_{\text{B}}$ and the contrastive loss $\mathcal{L}_{\text{C}}$ for Back-Propagation (BP). \label{fig:correspondence_arch}}
    \end{figure}

	Until this stage, it is premised that ellipsoid radii $\mathbf{r}_n$ are drawn from detected echoes $t_{n,k}^\star$ originating from the same distinct target. However, real-world scenes comprise complex topologies with reflections from multiple objects resulting in several echoes per channel. In particular, echoes emanating from different targets yield false positive solutions $\mathbf{s}^\star$. Hence, inter-channel echo assignment is a crucial, non-trivial undertaking for the proposed echolocation scheme to work and addressed hereafter. An overview of the architectural design for the echo association is outlined in Fig.~\ref{fig:correspondence_arch}.
	
	\subsection{Echo Feature Extraction}
    The extraction of acoustic features has been widely explored using Generalized Cross-Correlation (GCC) methods, for instance, Phase Transform (PhaT)~\cite{padois2019use, lee2020, Tracy:2021}. For reasons of phase invariance, we employ the \mbox{MEMG} model~\cite{Hahne:2022} instead as a viable starting point for reducing echo information while the oscillation term is skipped here. %
	Accordingly, an echo is modeled as $m(\mathbf{p};t_i)$ given by
	\begin{align}
	m(\mathbf{p};t_i) = \alpha\exp\left(-\frac{\left(t_i-\mu\right)^2}{2\sigma^2}\right) \left(1+\text{erf}\left(\eta\frac{t_i-\mu}{\sigma\sqrt{2}}\right)\right)
    \label{eq:memg_model}
	\end{align}
	with parameters \mbox{$\mathbf{p}=\left[\alpha,\mu,\sigma,\eta\right]^\intercal \in \mathbb{R}^{4 \times 1}$}. The $\exp(\cdot)$ term is the uni-variate Gaussian distribution with $\mu$ as the mean and $\sigma$ being the standard deviation. While $\alpha$ controls the echo amplitude, the exponentially modified term covers asymmetric shapes with $\eta$ and $\text{erf}(\cdot)$ as the error function. Integrating over $k\in\{1, 2, \dots, K\}$ echo components takes care of the multimodal distribution $M\left(\mathbf{\hat{p}}_n;t_i\right)$, which reads
	\begin{align}
	M\left(\mathbf{\hat{p}}_n;t_i\right)=\sum_{k=1}^K \, m(\mathbf{p}_k;t_i)
    \label{eq:memg_sum}
	\end{align}
	where frame vector \mbox{$\mathbf{\hat{p}}_n=\left[\mathbf{p}_1^\intercal, \mathbf{p}_2^\intercal, \mathbf{p}_k^\intercal, \dots, \mathbf{p}_K^\intercal\right]^\intercal\in\mathbb{R}^{4K}$} concatenates each echo variable $\mathbf{p}_k$ from frame $n$. 
	Each $\mathbf{\hat{p}}_n^\star$ is estimated using an optimization framework to minimize the energy $L(\mathbf{\hat{p}}_n)$ by
	\begin{align}
	L(\mathbf{\hat{p}}_n) = \left\lVert y_n(t_i)-M\left(\mathbf{\hat{p}}_n;t_i\right)\right\rVert^2_2
    \label{eq:memg_loss}
	\end{align}
	for every frame $n$. The Levenberg-Marquardt solver is used for minimization of \eqref{eq:memg_loss} where Hessians are obtained from analytical Jacobians \mbox{w.r.t.} $\mathbf{\hat{p}}^{(j)}_n$ at each iteration $j$. The best approximated MEMG vector $\mathbf{\hat{p}}^\star_n$ is given by
	\begin{align}
	\mathbf{\hat{p}}^\star_n=\underset{\mathbf{\hat{p}}_n^{(j)}}{\operatorname{arg\,min}} \, \left\{L\left(\mathbf{\hat{p}}^{(j)}_n\right)\right\} 
    \label{eq:memg_goal}
	\end{align}
	which carries echo component estimates $\mathbf{p}^{\star}_{n,k}\in\mathbf{\hat{p}}_n^\star$. For more details on robust MEMG convergence, the interested reader is referred to the original paper~\cite{Hahne:2022}. As in~\cite{Hahne:2022}, $\mathbf{p}^{\star}_{n,k}$ are extended by the hand-crafted frame confidence $C_n$, echo confidence $c_{n,k}$ as well as ToA $t^\star_{n,k}$ and echo power $p_{n,k}$, which is obtained by
	\begin{align}
	p_{n,k}=\sum_{i=1}^T m(\mathbf{p}^\star_{n,k}; t_i) \, , \quad \forall n, k
    \label{eq:memg_power}
	\end{align}
	so that \mbox{$\mathbf{\tilde{p}}^{\star}_{n,k}=[{\mathbf{p}^{\star}_{n,k}}^\intercal, c_{n,k}, p_{n,k}, t^\star_{n,k}, C_n]\in\mathbb{R}^{1\times8}$} serves as the input for the subsequent echo association.

	\subsection{Feature Correspondence}
	According to the Siamese correspondence architecture outlined in Fig.~\ref{fig:correspondence_arch}, echo features $\mathbf{\tilde{p}}^{\star}_{n,k}$ are fed into a Multi-Layer Perceptron (MLP) for echo selection and correspondence decision-making. The scalar output of each MLP reads
	\begin{align}
		b^{(n)}_k = h_4(h_3(h_2(h_1(\mathbf{\tilde{p}}^{\star}_{n,k})))) \, , \quad \forall n,k
	\end{align}
	where $h_l(\cdot)$ denote MLP function layers at indices \mbox{$l\in\{1,2,3,4\}$}. Each layer $h_l(\cdot)$ is equipped with trainable weights $\mathbf{W}_l$ and activated by a Rectifier Linear Unit (ReLU) except for $h_4(\cdot)$, which is followed by the sigmoid function. Learnable weight dimensions correspond to \mbox{$\mathbf{W}_1\in\mathbb{R}^{8\times32}$}, \mbox{$\mathbf{W}_2\in\mathbb{R}^{32\times32}$}, \mbox{$\mathbf{W}_3\in\mathbb{R}^{32\times4}$} and \mbox{$\mathbf{W}_4\in\mathbb{R}^{4\times1}$} with respective bias weights. The Binary Cross Entropy (BCE) is employed to learn predictions $b_k$ during training via
	\begin{align}
	\mathcal{L}_{\text{B}}(Y_k,b_k) = \sum_{k=1}^{K}-\left(Y_k\log(b_k)+(1-Y_k)\log(1-b_k)\right)
	\label{eq:loss_bce}
	\end{align}
	where $Y_k\in\{0,1\}$ represents ground-truth binary labels for each echo $k$ and channel index $n$ while the latter is omitted in loss functions for the sake of readability. The BCE loss helps classify an appropriate reference echo $\mathbf{\tilde{p}}^\star_r$ via
	\begin{align}
	\mathbf{\tilde{p}}^\star_r = \underset{\mathbf{\tilde{p}}^{\star}_{n,k}}{\operatorname{arg\,min}} \,\,
	\{h_4(h_3(h_2(h_1(\mathbf{\tilde{p}}^{\star}_{n,k}))))\}
	\end{align}
    across all channels $n$ and echoes $k$. The sought echo correspondence is established through a dissimilarity score $d^{(n)}_k$ between learned Siamese feature layer embeddings given by
	\begin{align}
	d^{(n)}_k = \left\lVert h_3(h_2(h_1(\mathbf{\tilde{p}}^{\star}_r))) - h_3(h_2(h_1(\mathbf{\tilde{p}}^{\star}_{n,k}))) \right\rVert_2 \, , \,\, \forall n,k \label{eq:euclidean_distance}
	\end{align}
	where $\lVert\cdot\rVert_2$ computes the component-wise Euclidean distance. The dissimilarity $d_k$ indicates how reliably a selected echo matches the reference $\mathbf{\tilde{p}}^\star_r$. This similarity metric is used during training through the contrastive loss initially postulated by Hadsell~\textit{et al.}\cite{Hadsell:06} and given as
	\begin{align}
	\mathcal{L}_{\text{C}}(Y_k,d_k) = \sum_k^K \frac{(1-Y_k){d_k}^2}{2} + \frac{Y_k\left\{\max\{0, q-{d_k}^2\}\right\}}{2}
	\label{eq:contrastive}
	\end{align}
	for all $n$ where $q>0$ is the margin regulating the border radius.
	For training, the total loss is 
	\begin{align}
	\mathcal{L}_{\text{T}}(Y_k,b_k,d_k) = \lambda_{\text{C}} \mathcal{L}_{\text{C}}(Y_k,d_k) + \lambda_{\text{B}} \mathcal{L}_{\text{B}}(Y_k,b_k)
	\label{eq:loss}
	\end{align}
	where weights $\lambda_{\text{C}}$ and $\lambda_{\text{B}}$ determine the loss ratio. 
	\section{Experimental Work}%
	\label{sec:experiment}
	\begin{figure*}[!b]
		\centering
		\begin{minipage}{.215\textwidth}
			\includegraphics[width=\textwidth]{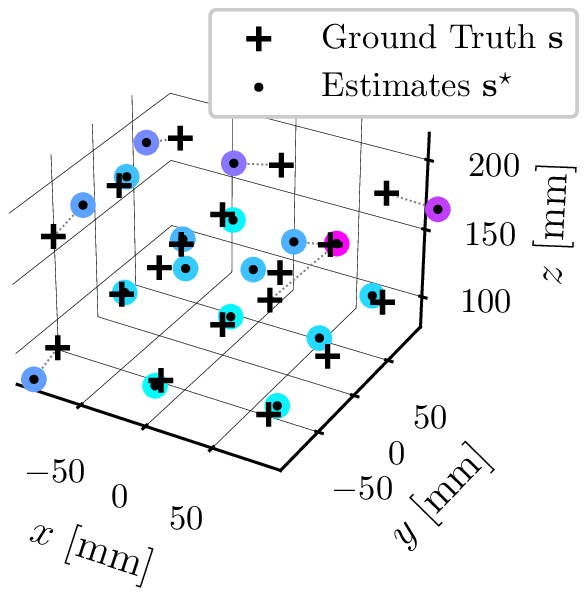}
		\end{minipage}
		\hfill
		\begin{minipage}{.215\textwidth}
			\includegraphics[width=\textwidth]{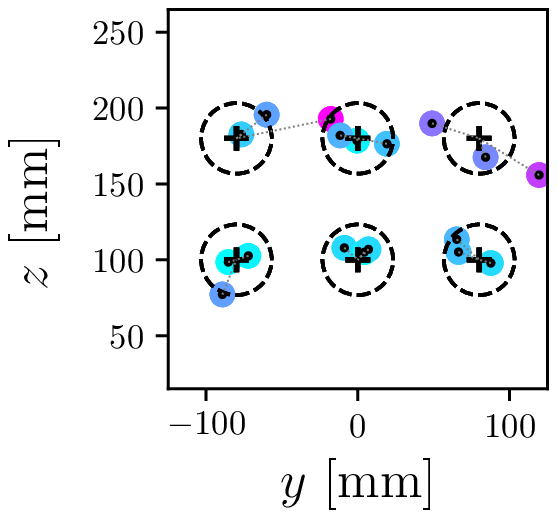}
		\end{minipage}
		\hfill
		\begin{minipage}{.215\textwidth}
			\includegraphics[width=\textwidth]{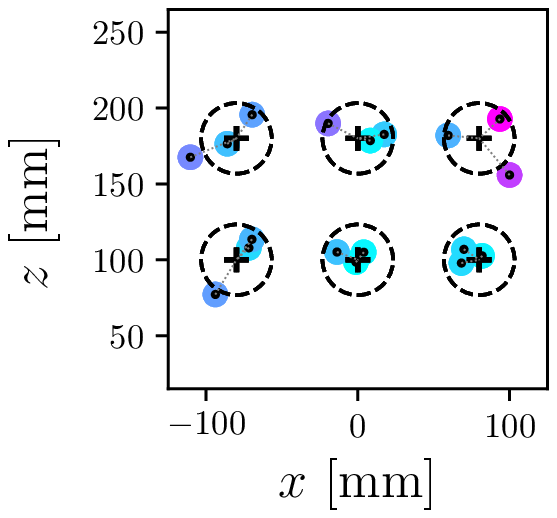}
		\end{minipage}
		\hfill
		\begin{minipage}{.215\textwidth}
			\includegraphics[width=\textwidth]{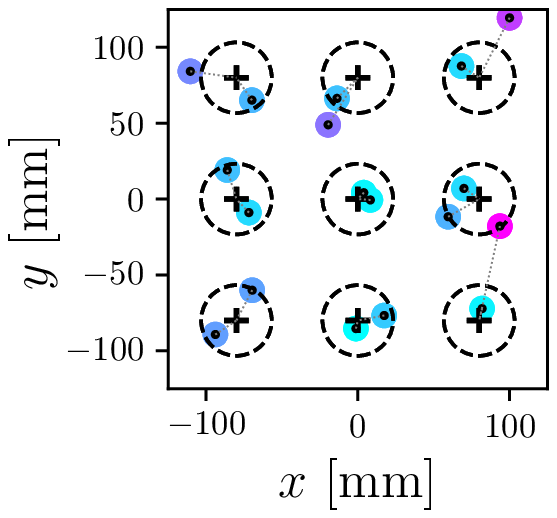}
		\end{minipage}
		\hfill
		\begin{minipage}{.095\textwidth}
			\includegraphics[width=\textwidth]{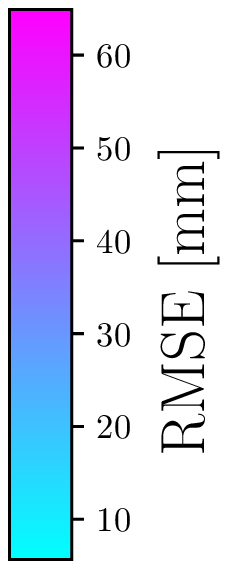}
			\vspace{.15pt}
		\end{minipage}
		\caption{\textbf{Experimental object localization results} showing 18 position estimates $\mathbf{s}^\star$ in the $[-80~\si{\milli\meter}, 0~\si{\milli\meter}, 80~\si{\milli\meter}]$ interval of the ($xy$)-plane and $[100~\si{\milli\meter}, 180~\si{\milli\meter}]$ interval in $z$-direction. The left diagram shows $\mathbf{s}^\star$ in 3-D, whereas adjacent plots depict 2-D projections of the same. Colors illustrate the individual RMSE while dashed circles represent the mean RMSE.\label{fig:space_results}}
	\end{figure*}
	%
	\subsection{Prototyping}
\begin{figure}[H]
		\centering
		\includegraphics[width=\linewidth]{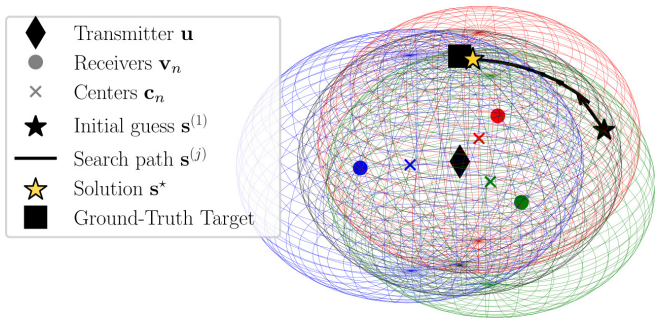}
		\caption{\textbf{Localization result based on ellipsoid intersection} showing the solution $\mathbf{s}^\star$, ground-truth position, transmitter $\mathbf{u}$ and receivers $\mathbf{v}_n$ with respective ellipsoids spanned by ToAs.\label{fig:ellipsoid_result}}
	\end{figure}

	A prototype sensor device is built from $N=4$ airborne Micro-Electro-Mechanical Systems (MEMS) transducers offering low power consumption and a compact form factor. 
	Pulse emittance and echo reception benefit from a \mbox{180\si{\degree}-wide} field-of-view that enables omni-directional tracking. The transducers operate at 175~\si{\kilo\hertz}, whereas each receiver captures \mbox{$T=64$} samples at a frequency of 22~\si{\kilo\hertz}. 
	Figure~\ref{fig:ellipsoid_result} depicts an ellipsoid intersection where receiver positions span an equilateral triangle with the emitter located at its centroid having a point-symmetry in mind. The spacing between each transducer and the emitter is a radius \mbox{$B=75~\si{\milli\meter}$}. 
	%
	\subsection{Dataset and Training}
	For data acquisition, a six-axis, vertically-articulated robot arm (Meca500) is employed to navigate a convex target to Ground-Truth (GT) positions (see Fig.~\ref{fig:top}). To suppress reflections from the robot, frames are subtracted by captures from an empty run in the absence of the target. 
	A dedicated training and validation set of 302 frames is captured by \mbox{$N=4$} sensors where each acquisition contains at least 3 detected echoes per channel, yielding approximately $3600$ EMG components for training overall. From this, a fraction of 0.3 is reserved as a validation set. Labels $Y_k$ are inferred by projecting GT positions as GT ToA $\mu_{gt}$ in the time domain. %
	Only a single MLP is trained due to Siamese networks sharing weights. Using an Adam optimizer, a learning rate of \num{5e-4} has shown to perform best. The frame batch size is 1, whereas losses of every $k$-th echo are back-propagated at each step. Weights are chosen to be $\lambda_{\text{C}}=1$ and $\lambda_{\text{B}}=10$ to balance the numerical loss gap. To prevent over-fitting, the maximum number of epochs is limited by early stopping criteria with $\text{tolerance} = 5$ and $\text{min. delta} = 0$. 
	%
	\subsection{Quantitative Results}
	%
	Numerical object localization results from the acquired test data taken with a robot from Fig.~\ref{fig:top} are provided in Table~\ref{tab:experimental}. \par%
	\begin{table}[!ht]
		\centering
		\caption{Experimental \mbox{3-D} localization results in \si{\milli\meter}}
		\label{tab:experimental}
		\begin{tabular}{rrr|rrr|rr}
			\multicolumn{3}{c|}{Ground-Truth $\mathbf{s}$} & \multicolumn{3}{c|}{Estimates $\mathbf{s}^\star$} & \multicolumn{2}{c}{RMSE} \\
			$x$ & $y$ & $z$ & $x^\star$ & $y^\star$ & $z^\star$ & [mm] & [\%] \\
			\hline\hline
            -80.0 & -80.0 & 100.0 & -93.9 & -89.2 & 77.1 & 28.3 & 18.7 \\
            -80.0 & -80.0 & 180.0 & -69.6 & -60.1 & 195.6 & 27.4 & 12.9 \\
            -80.0 & 0.0 & 100.0 & -71.8 & -8.9 & 107.9 & 14.4 & 11.3 \\
            -80.0 & 0.0 & 180.0 & -86.0 & 19.1 & 176.5 & 20.4 & 10.3 \\
            -80.0 & 80.0 & 100.0 & -69.8 & 65.2 & 113.5 & 22.5 & 14.9 \\
            -80.0 & 80.0 & 180.0 & -110.4 & 84.2 & 167.6 & 33.1 & 15.6 \\
            0.0 & -80.0 & 100.0 & -1.2 & -85.3 & 98.5 & 5.7 & 4.4 \\
            0.0 & -80.0 & 180.0 & 17.4 & -76.7 & 182.6 & 17.9 & 9.1 \\
            0.0 & 0.0 & 100.0 & 3.9 & 4.2 & 105.0 & 7.6 & 7.6 \\
            0.0 & 0.0 & 180.0 & 8.2 & -0.6 & 178.6 & 8.4 & 4.7 \\
            0.0 & 80.0 & 100.0 & -13.7 & 66.4 & 105.1 & 20.0 & 15.6 \\
            0.0 & 80.0 & 180.0 & -19.6 & 49.0 & 189.9 & 38.0 & 19.3 \\
            80.0 & -80.0 & 100.0 & 81.9 & -72.2 & 102.6 & 8.5 & 5.6 \\
            80.0 & -80.0 & 180.0 & 93.6 & -17.8 & 192.8 & 64.9 & 30.5 \\
            80.0 & 0.0 & 100.0 & 69.9 & 7.0 & 106.9 & 14.1 & 11.0 \\
            80.0 & 0.0 & 180.0 & 59.6 & -11.6 & 182.0 & 23.6 & 12.0 \\
            80.0 & 80.0 & 100.0 & 68.3 & 87.7 & 97.8 & 14.2 & 9.4 \\
            80.0 & 80.0 & 180.0 & 100.0 & 119.4 & 155.9 & 50.4 & 23.7 \\
			\hline\hline
			\multicolumn{6}{r|}{Mean} & $23.3$ & $13.1$ \\
			\multicolumn{6}{r|}{Std.} & $15.1$ & $6.6$ \\
		\end{tabular}
	\end{table}
	An important observation from this experiment is a tendency of more significant errors with increasing radial distance from the \mbox{$xy$-origin} $\mathbf{u}=[0,0,z]^\intercal$ of the sensor device. To make this visible, Fig.~\ref{fig:space_results} depicts cross-sectional projections of the results. 
    The radial error in $(x,y)$ is expected as minor deviations in relative ToAs (i.e., TDoAs) produce an enormous impact when jointly propagating to the \mbox{$xy$-plane} during an ellipsoid intersection. %
	Therefore, the prototype sensor arrangement is radially symmetric, creating a point-symmetric error distribution around the centre $\mathbf{u}$ in the ideal case. It is also essential to consider that deviations are specific to the sensor hardware. This includes a limited temporal resolution \mbox{($T=64$)}, just as potential mechanical sensor misalignments. Also, minor object movements caused sonic interference in the experiment, letting target echoes fluctuate and almost disappear in certain frames. Besides that, echo detection and MEMG regression occasionally fuse closely overlapping echoes to a single detected component. \par%
    %
    To set the results from Table~\ref{tab:experimental} in a broader context, a fair comparison of the proposed model with state-of-the-art methods would involve using the same sensor hardware and setup, which exceeds the scope of this study. Instead, error measures from closely related experiments are reported hereafter. Recently, Juan and Hu~\cite{Chung-Wei:21} presented a 2-D finger position tracking with an RMSE of $0.7\pm0.5$~cm using an extended Kalman filter for 6 transducers, each running at 40~kHz. The 3-D object tracking device released by manufacturer Toposens~\cite{Toposens:2022} achieves $1.0\pm2.5$~cm errors by correlating bounced-off phase signals from 3 receiving transducers running at 40~kHz, which are perpendicularly placed to each other in the range of the wavelength. Given that these deviations are from different sensor hardware, the results in Table~\ref{tab:experimental} are within the expected error range. %
    \subsection{Echo association}
\begin{figure}[!ht]
		\centering
		\includegraphics[width=\linewidth]{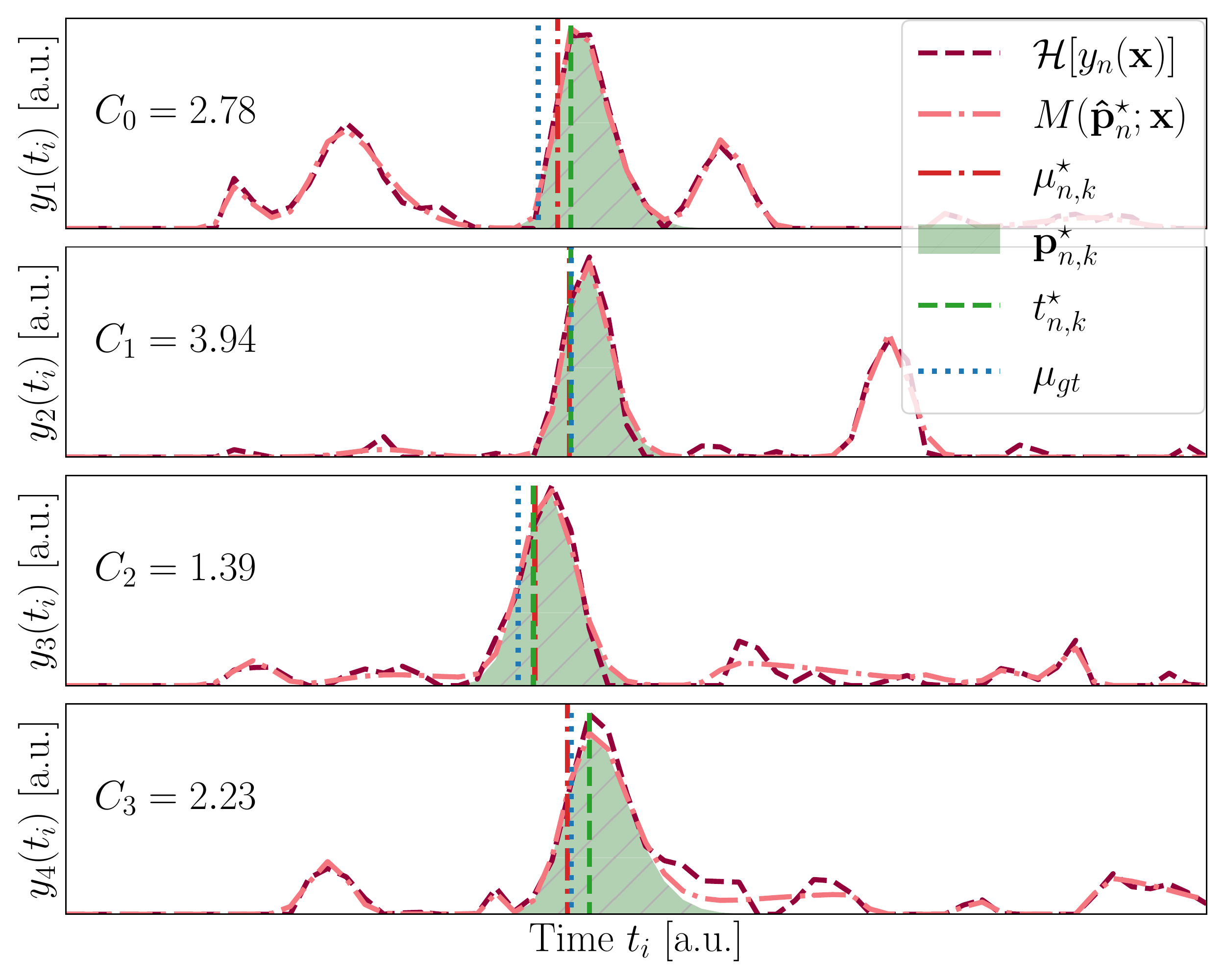}
		\caption{\textbf{Echo correspondence in A-Scan frames} $y_n(t_i)$ at GT position $\mathbf{s}=\left[\mbox{80.0~\si{\milli\meter}}, \mbox{0.0~\si{\milli\meter}}, \mbox{180.0~\si{\milli\meter}}\right]$ captured by $N=4$ sensors with echo association highlighted in green as hatched EMG components $\mathbf{p}_{n,k}^\star$ and frame confidence $C_n$ as defined in~\cite{Hahne:2022}. See the resolved echo ambiguity in $y_1(t_i)$.\label{fig:a_scan}}
	\end{figure}

 	Another crucial premise for PaCE to work is the proposed echo correspondence solver, which gives a promising \mbox{$F_1$-score} of 1.0 on the test data from Table~\ref{tab:experimental}, where all 18 echo correspondences are matched correctly. Figure~\ref{fig:a_scan} depicts an exemplary correspondence data sample. \par
    Table~\ref{tab:ablation} provides an overview of each module's impact on the overall matching performance of the proposed framework. %
    \begin{table}[!ht]
        \vspace{1.25mm}
        \centering
        \caption{Ablation overview for echo association}
        \label{tab:ablation}
        \begin{tabular}{l|c|c|c|c}
            Features & MEMG & MEMG & $[t^{\star}_{k}, \alpha^{\star}_k]$ & MEMG\\
            \hline
            Reference & $\operatorname{arg\,max}(\alpha_k)$ & MLP & MLP & MLP \\ 
            \hline
            Association & Munkres & Munkres & Contrastive & Contrastive \\
            \hline\hline
            Accuracy & 0.7778 & 0.8889 & 0.8889 & 1.0000 \\
            $F_1$-score & 0.5173 & 0.4682 & 0.9866 & 1.0000 \\
        \end{tabular}
	\end{table}
    The ablation study of the echo correspondence network is carried out by substituting the MLP from \eqref{eq:loss_bce} with an $\operatorname{arg\,max}$ operator of the amplitude scale $\alpha_k$ and the contrastive loss from \eqref{eq:contrastive} with the Munkres (also known as Hungarian) algorithm. The impact of MEMG features from \cref{eq:memg_model,eq:memg_sum,eq:memg_loss,eq:memg_goal,eq:memg_power} is evaluated by its replacement with the ToA $t^{\star}_k$ from \eqref{eq:distance} and its amplitude value $\alpha^{\star}_k$. Table~\ref{tab:ablation} demonstrates that MLP and contrastive loss outperform the \mbox{Munkres-based} echo association accuracy. Furthermore, MEMG features are more reliable correspondence indicators than just using ToAs. A suitable alternative to MEMG, e.g., based on learned convolutions, is yet to be devised since existing methods in the field (e.g., PhaT~\cite{padois2019use, lee2020}) employ waveform data. 
	%
	\section{Conclusions}
	\label{sec:conclusions}
	%
	With robots manoeuvring in 3-D space, tomorrow's computational vision requires reliable recognition of surroundings under difficult lighting conditions. %
	Through quantitative assessment of actual targets, this study demonstrates that the proposed PaCE model facilitates \mbox{3-D} tracking by at least 3 arbitrarily located ToF detectors and one emitter without phase information. A broad variance of RMSEs is reported by relative ToA displacements owing to sensor characteristics and occasional correspondence mismatches. \par%
	Once PaCE reaches maturity, it will serve as a considerable and cost-effective alternative to beamforming, joining the ranks of Delay-And-Sum (DAS), Synthetic Aperture Focusing Technique (SAFT), and Direction-of-Arrival (DoA) algorithms. Although the results presented here are not yet comparable to the former, more established principles, this feasibility study lays the groundwork for more active research to come. \par
	Follow-up studies will further investigate the performance of PaCE in an extensive experimental analysis, for instance, when ported to other sensor hardware. Its ability to localize several objects simultaneously will be an essential milestone. Exploiting the temporal domain via target tracking will stabilize the proposed localization scheme. Another central research question will be how PaCE can support SLAM as part of a multi-modal data fusion scheme. %

	\addtolength{\textheight}{-9.6cm}

	\section*{ACKNOWLEDGMENT}
	
	The author hereby thanks Milica Bulatovic for kindly sharing the laboratory equipment as well as Meret Ruch and Urs Rohrer for helping with the prototype's mechanical design and assembly. The author is also grateful to Raphael Sznitman for his invaluable advice throughout this research project at the ARTORG Center. This project is funded by the Hasler Foundation under number 22027 and the author's gratitude is also extended to the foundation for their trust and support.
	%
	
	
	\bibliographystyle{IEEEtran}
	\bibliography{root}

\begin{thebibliography}{10}
\providecommand{\url}[1]{#1}
\csname url@samestyle\endcsname
\providecommand{\newblock}{\relax}
\providecommand{\bibinfo}[2]{#2}
\providecommand{\BIBentrySTDinterwordspacing}{\spaceskip=0pt\relax}
\providecommand{\BIBentryALTinterwordstretchfactor}{4}
\providecommand{\BIBentryALTinterwordspacing}{\spaceskip=\fontdimen2\font plus
\BIBentryALTinterwordstretchfactor\fontdimen3\font minus
  \fontdimen4\font\relax}
\providecommand{\BIBforeignlanguage}[2]{{%
\expandafter\ifx\csname l@#1\endcsname\relax
\typeout{** WARNING: IEEEtran.bst: No hyphenation pattern has been}%
\typeout{** loaded for the language `#1'. Using the pattern for}%
\typeout{** the default language instead.}%
\else
\language=\csname l@#1\endcsname
\fi
#2}}
\providecommand{\BIBdecl}{\relax}
\BIBdecl

\bibitem{Peremans:1993}
H.~Peremans, K.~Audenaert, and J.~Van~Campenhout, ``A high-resolution sensor
  based on tri-aural perception,'' \emph{IEEE Transactions on Robotics and
  Automation}, vol.~9, no.~1, pp. 36--48, 1993.

\bibitem{Wijk:2000}
O.~Wijk and H.~Christensen, ``Triangulation-based fusion of sonar data with
  application in robot pose tracking,'' \emph{IEEE Transactions on Robotics and
  Automation}, vol.~16, no.~6, pp. 740--752, 2000.

\bibitem{Bank:2007}
D.~Bank and T.~Kampke, ``High-resolution ultrasonic environment imaging,''
  \emph{IEEE Transactions on Robotics}, vol.~23, no.~2, pp. 370--381, 2007.

\bibitem{Mitsubishi:2016}
K.~Araki, R.~Suzuki, S.~Inoue, and Y.~Nishimoto, ``Obstacle detection device
  and obstacle detection method,'' Mitsubishi Electric Corp., Patent No.
  JP2016128769A, Jul 2016.

\bibitem{Toposens:2016}
A.~Rudoy, ``3d-position determination method and device,'' Toposens GmbH,
  Patent No. US10698094B2, Aug 2016.

\bibitem{Valeo:2019}
A.~Walz and F.~Thunert, ``Method for detecting an object in an environmental
  region of a motor vehicle by means of an ultrasound sensor device by
  determining a {3-D} position of an object point, ultrasound sensor device and
  driver assistance system,'' Valeo Schalter und Sensoren GmbH, Patent No.
  DE102018102350A1, Aug 2019.

\bibitem{Reijniers:2007}
J.~Reijniers and H.~Peremans, ``Biomimetic sonar system performing
  spectrum-based localization,'' \emph{IEEE Transactions on Robotics}, vol.~23,
  no.~6, pp. 1151--1159, 2007.

\bibitem{Schillebeeckx:2011}
F.~Schillebeeckx, F.~D. Mey, D.~Vanderelst, and H.~Peremans, ``Biomimetic
  sonar: Binaural 3d localization using artificial bat pinnae,'' \emph{The
  International Journal of Robotics Research}, vol.~30, no.~8, pp. 975--987,
  2011.

\bibitem{Eliakim:2018}
\BIBentryALTinterwordspacing
I.~Eliakim, Z.~Cohen, G.~Kosa, and Y.~Yovel, ``A fully autonomous terrestrial
  bat-like acoustic robot,'' \emph{PLOS Computational Biology}, vol.~14, no.~9,
  pp. 1--13, 09 2018. [Online]. Available:
  \url{https://doi.org/10.1371/journal.pcbi.1006406}
\BIBentrySTDinterwordspacing

\bibitem{Christensen:2020}
J.~H. Christensen, S.~Hornauer, and S.~X. Yu, ``Batvision: Learning to see 3d
  spatial layout with two ears,'' in \emph{2020 IEEE International Conference
  on Robotics and Automation (ICRA)}, 2020, pp. 1581--1587.

\bibitem{Tracy:2021}
E.~Tracy and N.~Kottege, ``{CatChatter}: Acoustic perception for mobile
  robots,'' \emph{IEEE Robotics and Automation Letters}, vol.~6, no.~4, pp.
  7209--7216, 2021.

\bibitem{Parida:2021}
K.~K. Parida, S.~Srivastava, and G.~Sharma, ``Beyond image to depth: Improving
  depth prediction using echoes,'' in \emph{Proceedings of the IEEE/CVF
  Conference on Computer Vision and Pattern Recognition}, 2021, pp. 8268--8277.

\bibitem{Irie:22}
G.~Irie, T.~Shibata, and A.~Kimura, ``Co-attention-guided bilinear model for
  echo-based depth estimation,'' in \emph{ICASSP 2022 - 2022 IEEE International
  Conference on Acoustics, Speech and Signal Processing (ICASSP)}, 2022, pp.
  4648--4652.

\bibitem{pace:2022}
C.~Hahne and R.~Sznitman, ``Parallax among corresponding echoes,'' Patent No.
  EP22197595.6, Sep 2022.

\bibitem{Ahmad:2006}
F.~Ahmad and M.~G. Amin, ``Noncoherent approach to through-the-wall radar
  localization,'' \emph{IEEE Transactions on Aerospace and Electronic Systems},
  vol.~42, no.~4, pp. 1405--1419, 2006.

\bibitem{Mirbach:2011}
M.~Mirbach, ``A simple surface estimation algorithm for uwb pulse radars based
  on trilateration,'' in \emph{2011 IEEE International Conference on
  Ultra-Wideband (ICUWB)}, 2011, pp. 273--277.

\bibitem{malanowski:2012}
M.~Malanowski and K.~Kulpa, ``Two methods for target localization in
  multistatic passive radar,'' \emph{IEEE Transactions on Aerospace and
  Electronic Systems}, vol.~48, no.~1, pp. 572--580, 2012.

\bibitem{kuc:2016}
R.~Kuc and V.~Kuc, ``Modeling human echolocation of near-range targets with an
  audible sonar,'' \emph{The Journal of the Acoustical Society of America},
  vol. 139, no.~2, pp. 581--587, 2016.

\bibitem{Chung-Wei:21}
\BIBentryALTinterwordspacing
C.-W. Juan and J.-S. Hu, ``Object localization and tracking system using
  multiple ultrasonic sensors with newton-raphson optimization and kalman
  filtering techniques,'' \emph{Applied Sciences}, vol.~11, no.~23, 2021.
  [Online]. Available: \url{https://www.mdpi.com/2076-3417/11/23/11243}
\BIBentrySTDinterwordspacing

\bibitem{Haxthausen:2021}
F.~von Haxthausen, S.~B{\"o}ttger, D.~Wulff, J.~Hagenah, V.~García-Vázquez,
  and S.~Ipsen, ``Medical robotics for ultrasound imaging: Current systems and
  future trends,'' \emph{Current Robotics Reports}, vol.~2, no.~1, pp.
  2662--4087, 2021.

\bibitem{hahne2017baseline}
C.~Hahne, A.~Aggoun, V.~Velisavljevic, S.~Fiebig, and M.~Pesch, ``Baseline and
  triangulation geometry in a standard plenoptic camera,'' \emph{International
  Journal of Computer Vision}, pp. 1--15, 8 2017.

\bibitem{Hahne:2022}
C.~Hahne, ``Multimodal exponentially modified gaussian oscillators,'' in
  \emph{2022 IEEE International Ultrasonics Symposium (IUS)}, 2022, pp. 1--4.

\bibitem{padois2019use}
T.~Padois, O.~Doutres, and F.~Sgard, ``On the use of modified phase transform
  weighting functions for acoustic imaging with the generalized cross
  correlation,'' \emph{The Journal of the Acoustical Society of America}, vol.
  145, no.~3, pp. 1546--1555, 2019.

\bibitem{lee2020}
R.~Lee, M.-S. Kang, B.-H. Kim, K.-H. Park, S.~Q. Lee, and H.-M. Park, ``Sound
  source localization based on gcc-phat with diffuseness mask in noisy and
  reverberant environments,'' \emph{IEEE Access}, vol.~8, pp. 7373--7382, 2020.

\bibitem{Hadsell:06}
R.~Hadsell, S.~Chopra, and Y.~LeCun, ``Dimensionality reduction by learning an
  invariant mapping,'' in \emph{2006 IEEE Computer Society Conference on
  Computer Vision and Pattern Recognition (CVPR'06)}, vol.~2, 2006, pp.
  1735--1742.

\bibitem{Toposens:2022}
{Toposens GmbH}, ``{ECHO ONE}: 3d ultrasonic echolocation and ranging sensor,''
  Data Sheet V1.0, July 2022.

\end{thebibliography}
	
\end{document}